\documentclass[10pt,twocolumn,letterpaper]{article}

\usepackage{iccv}
\usepackage{times}
\usepackage{epsfig}
\usepackage{graphicx}
\usepackage{amsmath}
\usepackage{amssymb}

\usepackage{booktabs}
\usepackage{warpcol}
\usepackage{enumitem}
\usepackage{algorithm}
\usepackage{algpseudocode}

\usepackage[pagebackref=true,breaklinks=true,letterpaper=true,colorlinks,bookmarks=false]{hyperref}

\iccvfinalcopy 

\newcommand{\comments}[1]{\textcolor{black}{#1}}
\newcommand{\ke}[1]{{\color{black}#1}}

\newcommand{\jiayu}[1]{\textcolor{black}{#1}}

\ificcvfinal\fi

\begin{document}

\title{MODNet-V: Improving Portrait Video Matting via Background Restoration}

\author{
Jiayu Sun$^{1,2}$\hspace{-0.1em} \footnotemark[2]   \quad Zhanghan Ke$^{1}$ \hspace{-0.1em}\footnotemark[2]  \quad Lihe Zhang$^{2}$ \quad Huchuan Lu$^{2}$ \quad Rynson W.H. Lau$^{1}$\\
$^{1}$Department of Computer Science, City University of Hong Kong\\
$^{2}$School of Information and Communication Engineering,
Dalian University of Technology\\
}
\maketitle
\ificcvfinal\thispagestyle{empty}\fi
\renewcommand{\thefootnote}{\fnsymbol{footnote}}
\footnotetext[2]{Equal Contribution}
\footnotetext[1]{This manuscript is a preprint.}
\begin{abstract}
To address the challenging portrait video matting problem more precisely, existing works typically apply some matting priors that require additional user efforts to obtain, such as annotated trimaps or background images. In this work, we observe that instead of asking the user to explicitly provide a background image, we may recover it from the input video itself.
To this end, we first propose a novel background restoration module (BRM) to recover the background image dynamically from the input video. BRM is extremely lightweight and can be easily integrated into existing matting models. By combining BRM with a recent image matting model, MODNet, we then present MODNet-V for portrait video matting. Benefited from the strong background prior provided by BRM, MODNet-V has only 1/3 of the parameters of MODNet but achieves comparable or even better performances. 
Our design allows MODNet-V to be trained in an end-to-end manner on a single NVIDIA 3090 GPU.
Finally, we introduce a new patch refinement module (PRM) to adapt MODNet-V for high-resolution videos while keeping MODNet-V lightweight and fast. 
\end{abstract}

\section{Introduction}

Portrait video matting aims to extract the foreground portraits from a given video sequence. Accurately addressing this task in real time is essential for many practical applications, such as special effect generation in interactive video editing and background replacement in online meetings. Since video matting is a challenging problem due to the ill-posed nature of this task,
early works~\cite{chuang2002video,bai2011towards,zhou2016learning,sun2021deep} typically expect the user to provide a pixel-wise trimap for each frame or for several consecutive key frames, which is a useful prior with great matting results. However, this trimap is unavailable in real-time applications as it requires human annotation. To alleviate this problem, Background Matting~\cite{lin2021real} proposes \ke{to use a pre-captured background image as an additional input}. \ke{Although Background Matting achieves real-time performance, it} requires users to provide the additional background prior and will fail in scenarios with dynamic backgrounds (\ie, videos with changing background contents). Instead of requiring auxiliary matting priors, some recent methods~\cite{zhang2019late,qiao2020attention,liu2020boosting,ke2020green} attempt to predict a matte from just the input image. However, these methods often produce unsatisfactory alpha mattes in difficult cases, \ke{\textit{e.g.}, images with complex backgrounds}, since missing matting priors increases model learning difficulty. 
In this work, we observe that through dynamically accumulating as well as updating the background contents across consecutive video frames, we may recover a meaningful background image, which content changes dynamically as the background of the video changes, to serve as a prior for the matting task. 
Specifically, in a video sequence, the posture and location of the foreground portrait are usually changing continuously across frames, which means that the background contents in these frames are usually different and complementary. By accumulating and completing the missing background content frame-by-frame, we may produce a background image for the next frame. \ke{Compared with Background Matting~\cite{lin2021real}, our approach does not require any user efforts. In addition, if we allow the background content to be updated continuously, we may handle videos with dynamic backgrounds.}

Motivated by our observation, we propose a novel background restoration module (BRM) to help recover the image background dynamically.
BRM maintains two latent variables: (1) {\it background features} to represent the restored background, and (2) {\it a binary mask} to indicate the recovered background pixels. \ke{At each frame time, BRM performs two steps. First, it uses the background features from the current frame as a prior for the matting model to produce a predicted alpha matte. Second, it extracts the new background pixels from the current input frame and accumulates these background pixles into the background features. 
In this step, BRM will also update the existing background pixels, in order to handle the video sequences with dynamic backgrounds}. 
The design of BRM is extremely lightweight, with negligible computational overhead. BRM can be easily integrated into the existing matting models without modifying the original model architectures. 

\ke{If we apply background restoration as a matting prior, we may regard it as a sub-objective} of the portrait video matting objective in MODNet~\cite{ke2020green},  and we combine BRM with MODNet to form a new video matting model, named MODNet-V. Benefiting from the background prior provided by BRM, the proposed MODNet-V has only 1/3 of the parameters but achieves comparable or even better performance compared with MODNet. Moreover, we also introduce a new patch refinement module (PRM) to adapt MODNet-V to high-resolution videos
In summary, we make three-fold contributions. \ke{First, we observe that the background of a video frame can be restored from its previous frames, and we verify the feasibility of using the restored background as a video matting prior.} Second, we propose a novel background restoration module (BRM) to restore the background prior for video matting. Third, We present MODNet-V, a new light-weight model that contains the aforementioned BRM and a newly proposed patch refinement module (PRM). Our new model achieves notable improvements in model size, inference speed, and matting performance/stability.

\section{Related Work}
\subsection{Image Matting}

Matting is an inherently under-constrained problem as its formula contains three unknown variables ($F$, $B$, and $\alpha$) but only one known variable ($I$), as: 
\begin{equation}\label{eq:matting}
\begin{split}
I = \alpha F + (1-\alpha)B.
\end{split}
\end{equation}

 To solve Eq.\,(\ref{eq:matting}), most of existing matting methods take an auxiliary trimap as a prior. The earlier traditional methods can be classified into sampling-based methods~\cite{chuang2001bayesian,feng2016cluster,gastal2010shared,he2011global,johnson2016sparse,karacan2015image,ruzon2000alpha} and affinity-based methods~\cite{aksoy2018semantic,aksoy2017designing,bai2007geodesic,chen2013knn,grady2005random,levin2007closed,levin2008spectral}. However, these methods only  \jiayu{utilise} the low-level properties of image pixels. \jiayu{Instead, recent deep learning based methods leverage the high-level semantic information from neural networks and improve the matting results significantly.} For example, Cho et al.~\cite{cho2016natural} propose to combine neural network with close-form matting~\cite{levin2008spectral} and KNN matting~\cite{chen2013knn}. Xu et al.~\cite{xu2017deep} introduce the first end-to-end neural network for image matting. 
\jiayu{Besides, some methods promote the accuracy of alpha predictions by combining extra information (e.g., optimal trimaps~\cite{cai2019disentangled}, foreground information~\cite{hou2019context}, etc.) learned from an additional network.} 

However, annotating a trimap is costly and time-consuming in practice. To address this problem, some recent methods either directly predict the alpha matte from the input image~\cite{shen2016deep,levin2007closed,zhu2017fast,chen2018semantic,liu2020boosting,qiao2020attention,zhang2019late} or adopt the auxiliary prior~\cite{sengupta2020background}.
\jiayu{Among these methods, there are two different inputs for these methods: the natural images with diverse object categories and the portrait images.
In terms of the natural images with diverse object categories, Qiao et al.~\cite{qiao2020attention} apply attention mechanism and Zhang et al.~\cite{zhang2019late} integrate the possibility maps of foreground and background for alpha prediction. With regard to the portrait image matting,} Shen et al.~\cite{shen2016deep} apply a fully-convolutional network to generate a pseudo trimap before using an image matting layer to learn the Laplacian matrix. Chen et al.~\cite{chen2018semantic} first predict a low-resolution segmentation map, and then use it as matting guidance. Liu et al.~\cite{liu2020boosting} make use of coarsely annotated data in a three-stage coupled pipeline. Sengupta et al.~\cite{sengupta2020background} supply an extra background image as an alternative auxiliary cue to predict the alpha matte and foreground.
\vspace{2mm}
\begin{figure*}[t]
\centering
\begin{tabular}{c@{}c}
\hspace{-3mm}
\includegraphics[width=1\linewidth]{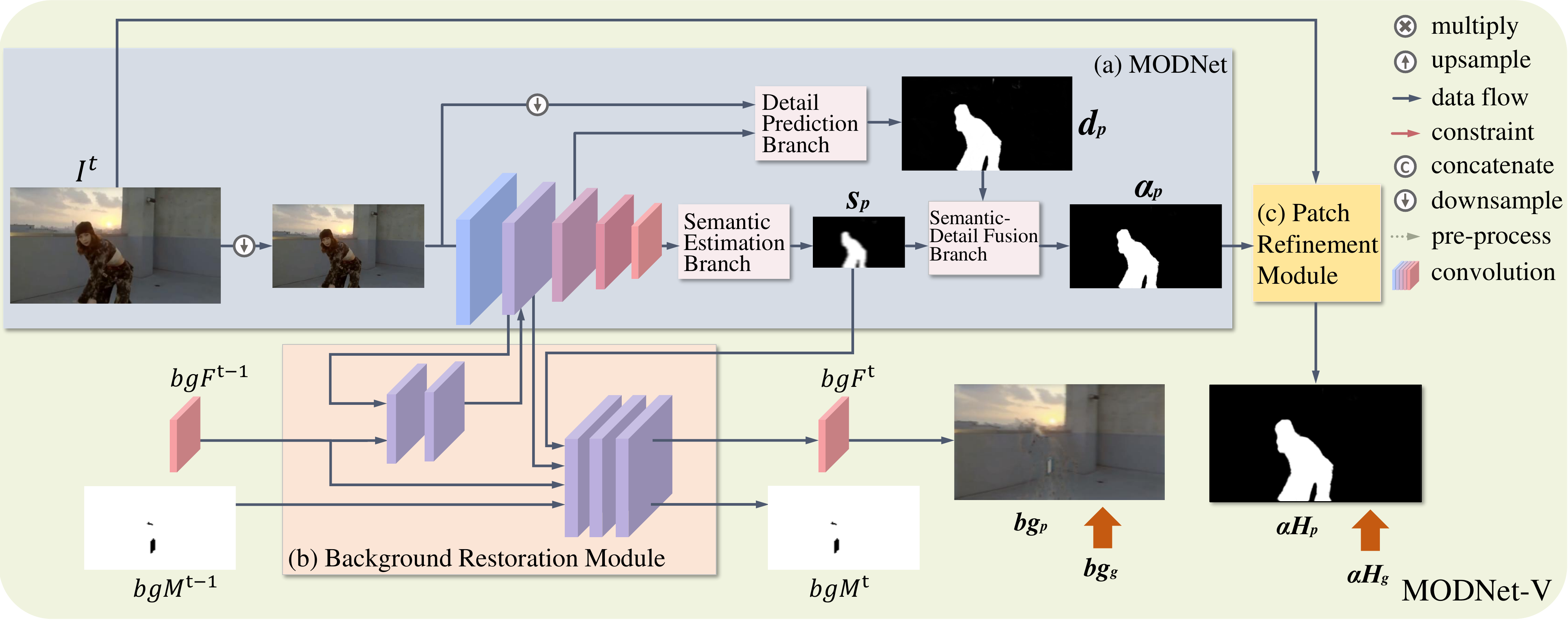}
\ \\
\end{tabular}
{\begin{center}
\vspace{-1mm}
\caption{\small{Architecture of MODNet-V. Our framework consists of a MODNet component adopted from \cite{ke2020green}, a background restoration module (BRM), and a patch refinement module (PRM). 
In each timestamp $t$, BRM provide a restored background feature $bgF^{t-1}$ as a matting prior and update $bgF$ under the guidance of a binary mask $bgM$. After the MODNet component predicts the alpha matte $\alpha_{p}$ from a downsample version of $I^t$, PRM detects the wrong patches in $\alpha_{p}$ and refines these patches in the original input resolution.}}
\label{fig:framework}
\end{center}
}
\vspace{-1mm}
\end{figure*}
\subsection{Video Matting}
Although we can simply apply aforementioned algorithms on per frame for video matting, it results in temporal incoherence. 
By taking temporal relationship of frames into consideration, previous trimap-based video matting methods~\cite{chuang2002video,bai2011towards,sun2021deep,zhou2016learning} created coherent trimaps or mattes of video sequence by performing spatio-temporal optimisations. For example, Chuang et al.~\cite{chuang2002video} propagate trimaps from a limited number of user-defined key-frames to the entire sequence by optical flow and predict alpha mattes via using Bayesian matting.  Bai et al.~\cite{bai2011towards} improve temporal coherence by a temporal matte filter while preserving matte structures on individual frames. \jiayu{Zou et al.~\cite{zhou2016learning} connect spatial and temporal relationship through non-local structure constructed by a sparse and low-rank representation. In addition to the methods employing low-level colour features, some learning-based methods employ deep structural and semantic features.} Some methods~\cite{choi2012video,li2013motion,zou2019unsupervised} utilise non-local matting Laplacian of multiple frames to encode temporal coherency. Moreover, Sun et al.~\cite{sun2021deep} introduce a trimap propagation network by spatio-temporal feature aggregation module. 

\jiayu{Recently, trimap-free video matting methods draw much attention.
Ke et al.~\cite{ke2020green} divide the matting objective into three sub-objectives and learn the consistency of each sub-objective for real-world portrait matting. However, this method fails in the coherence of prediction among frames as the temporal relationship has not been comprehensively considered.} Following the design of Sengupta et al.~\cite{sengupta2020background}, Lin et al.~\cite{lin2021real} also supply extra background image for real-time portrait video matting. 
This extra background is a strong prior for improving video matting performance, which has been demonstrated by the research of Lin et al~\cite{lin2021real}. However, using a captured background image as the prior has two main drawbacks which greatly limit the application of Background Matting in practice. \jiayu{First, background have to be static if using a captured background image, which means even small disturbances in the background (e.g., light changes and slight jitter) will affect the matting results.} Second, the background image requires additional user efforts to capture, and this process must be done carefully to ensure that the obtained background image is aligned with the video sequence. 
To eliminate the above issues, in this paper, we propose a novel BRM module, which restores background from the historical frames and then use it as a matting prior for coming frames.

\subsection{Scene background modeling}
Scene background modeling~\cite{bouwmans2017scene,petrosino2017scene} is the task of estimating a clear background image without foreground objects from a video sequence, which helps to identify the moving foreground objects in video. The scene background modeling methods always follow the same scheme.
\jiayu{They predict an unreliable background model using the first few frames as their first step, and then gradually update background model by the analysis of the extracted foreground objects from the video sequences in an online process.}
The most traditional background modeling approaches are based on Gaussian distribution~\cite{piccardi2004background,benezeth2008review}. Then, Stauffer and Grimson design a modeling of pixel-level color intensity variations based probability density functions which is a mixture of Gaussians~\cite{stauffer1999adaptive,stauffer2000learning}. 
\jiayu{Following the Gaussian mixture model, Zivkovic~\cite{zivkovic2004improved} introduces an adaptive Gaussian mixture model, which is an iterative approach between median blending and spatial segmentation. Some methods~\cite{lee2005effective,shimada2006dynamic} further improve the Gaussian mixture
model for background modeling. In addition, 
Kim et al.~\cite{kim2004background} establish the modeling of the background by a codebook method.}
Laugraud et al.~\cite{laugraud2017labgen} generate background based on motion detection. \jiayu{Javed et al.~\cite{javed2017background} formulate background–foreground modeling as a low-rank matrix decomposition problem through Robust Principal Component Analysis (RPCA). Recently, some deep-learning-based methods also emerge. These methods apply weightless neural network~\cite{de2017background}, convolutional neural network~\cite{halfaoui2016cnn,qu2016motion}, and auto-encoder~\cite{xu2014dynamic,xu2014motion} to the task of scene background modeling.}
The performance of the scene background modeling is profoundly affected by the presence of dynamic backgrounds, illumination changes, long-time static foreground, as well as global motions caused by camera jitter. 
In this paper, we try lessening the effects of dynamic backgrounds, illumination changes and other factors by the design of BRM.
\section{Method}
\begin{algorithm*}[t]
  \caption{The Proposed Background Restoration Method.}
  \label{alg::BG}
  \begin{small}
  \begin{algorithmic}[1]
    \Require 
        timestamp $t$;\;\;AND operation $\&$;\;\;pixel-by-pixel multiplication $\otimes$
    \Require
        background feature $bgF$;\;\;background binary mask$bgM$;\;\;background information $bgI$;\;\; predicted semantic $s_p$
    \State $bgM_{new}^{t} = (bgM^{t-1} == 0)\;\&\;((1 - s_p^t) > 0.5)$ \comments{// mask of background newly appeared in timestamp $bgI$}
    \State
    \State $bgM_{avg}^{t} = (bgM^{t-1} == 1)\;\&\;((1 - s_p^t) > 0.5)$ \comments{// mask of background appeared in both $bgF$ and $bgI$}
    \State
    \State $bgF^{t}_{tmp} = bgF^{t-1} + bgM_{new}^{t} \otimes bgI^{t}$ \comments{// add newly appeared background into $bgF$}
    \State
    \State $bgF^{t} = (1 - bgM_{avg}^{t}) \otimes bgF^{t}_{tmp} + bgM_{avg}^{t} \otimes ((bgF^{t}_{tmp} + bgI^{t})/2)$ \comments{// average background  appeared in both $bgF$ and $bgI$}
    \State
    \State $bgM^{t} = bgM^{t-1} + bgM_{new}^{t}$ \comments{// update $bgM$}
  \end{algorithmic}
  \end{small}
\end{algorithm*}
\subsection{Review of MODNet}
MODNet~\cite{ke2020green} proposed to decompose the challenging matting objective into three sub-objectives, including semantic estimation, detail prediction, and semantic-detail fusion. As shown in Fig.\,\ref{fig:framework} (a), in each timestamp $t$, MODNet (denoted by $f_{mod}$ here) predicts an alpha matte $\alpha_p^t$ from the input image frame $I^t$, as:
\begin{equation}\label{eq:modnet}
    \alpha_p^t = f_{mod}(I^t) = \mathcal{F}(s_p^t, d_p^t) = \mathcal{F}\Big(\mathcal{S}(I^t), \mathcal{D}\big(\mathcal{S}(I^t), I^t\big)\Big),
\end{equation}
where $\mathcal{S}$, $\mathcal{D}$, and $\mathcal{F}$ are the three branches in MODNet. $s_p^t$ and $d_p^t$ are the predicted semantic and detail respectively. However, Eq.\,(\ref{eq:modnet}) does not take any matting prior as the input and overlooks the temporal information, leading to unsatisfactory alpha mattes predicted under complex scenarios and inconsistent alpha mattes predicted among video frames. 

\subsection{Overview of MODNet-V}
To overcome the aforementioned problems of MODNet, we present MODNet-V, a light-weight video matting model contains newly proposed BRM and PRM.

BRM restores the background information from the historical frames. In MODNet-V, we use the background information restored by BRM as a matting prior, which is input to the semantic branch $\mathcal{S}$ along with the coming frame. Such a prior potentially contains the temporal information that is essential for video matting. We connect BRM to $\mathcal{S}$ by using:
\begin{equation}\label{eq:modnet-v-B}
    s_p^t = \mathcal{S}\big(\,I^t, \mathcal{B}(I^t)\,\big),
\end{equation}
where $\mathcal{B}$ denotes BRM.
Another module, PRM, refines the alpha matte predicted by Eq.\,(\ref{eq:modnet}) in a patch-based manner. As a result, MODNet-V can process high-resolution ({\it e.g.}, 1080p or 2k) images with much less computational overhead than MODNet. We append PRM as:
\begin{equation}\label{eq:modnet-v-P}
    \alpha_p^t = \mathcal{P}\big(\,\mathcal{F}(s_p^t, d_p^t)\,\big),
\end{equation}
where $\mathcal{P}$ denotes PRM. By combining Eq.\,(\ref{eq:modnet})(\ref{eq:modnet-v-B})(\ref{eq:modnet-v-P}), we can obtain the inference formula of MODNet-V:
\begin{equation}\label{eq:modnet-v}
    \alpha_p^t = \mathcal{P}\bigg(\mathcal{F}\Big(\,\mathcal{S}(I^t, \mathcal{B}(I^t)),\;\mathcal{D}\big(\,\mathcal{S}(I^t, \mathcal{B}(I^t)), I^t\,\big)\,\Big)\bigg).
\end{equation}

 \begin{figure}[t]
\centering
\begin{tabular}{c@{}c}
\hspace{-4mm}
\includegraphics[width=1\linewidth]{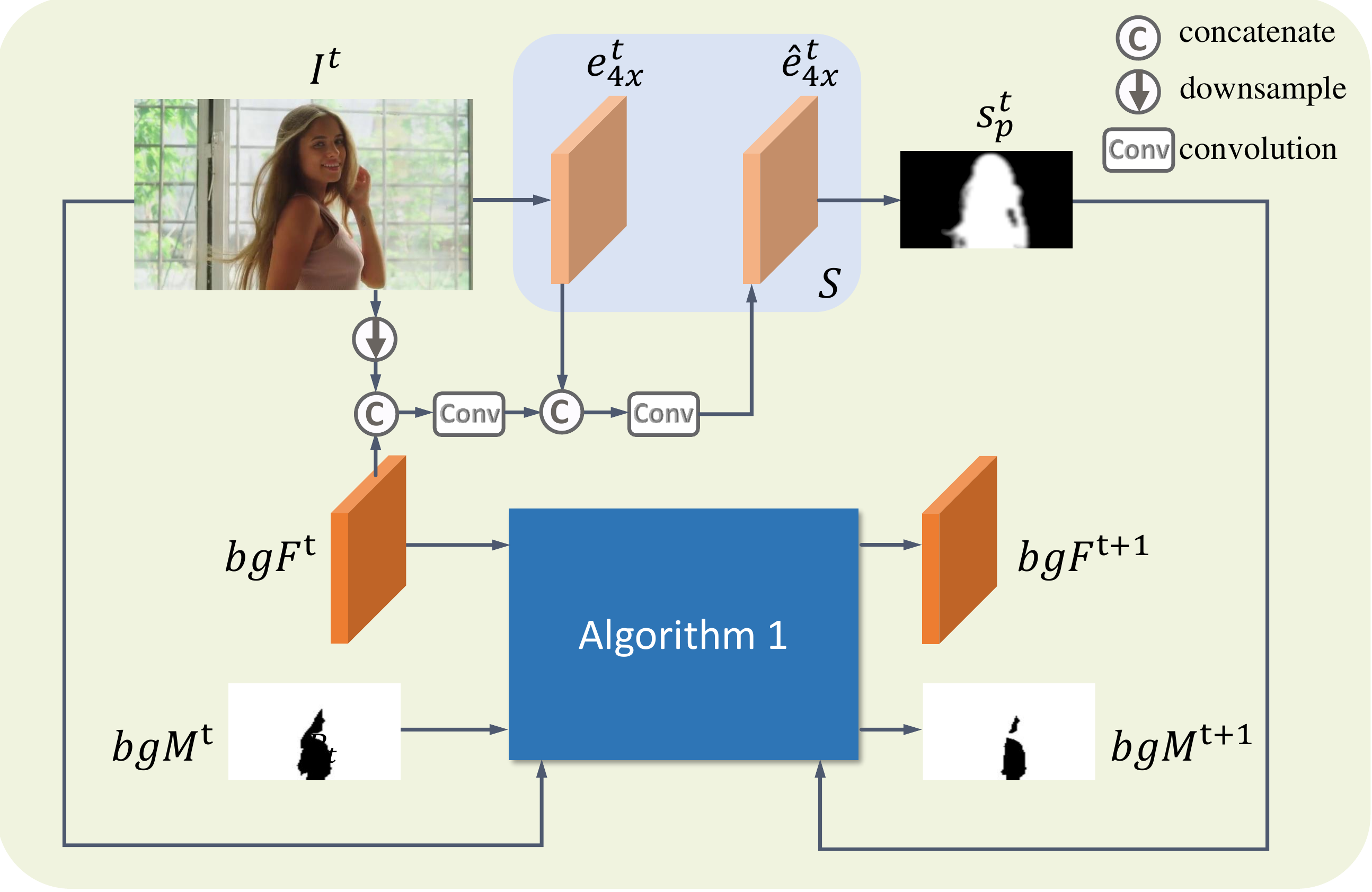}
\ \\
\end{tabular}
{\begin{center}
\vspace{-1mm}
\caption{\small{Illustration of BRM.}}
\label{fig:brm}
\end{center}
}
\vspace{-8mm}
\end{figure}

Benefit from BRM and PRM, we can also simplify the MODNet backbone used in MODNet-V to further reduce computational overhead. Specifically, we make three modifications: (1) reducing the input resolution of $\mathcal{S}$, (2) deleting the last encoder block (the 32x downsampling block), and (3) removing several convolutional layers in $\mathcal{D}$.

In the following two subsections, we will delve into BRM and PRM respectively.

\subsection{Background Restoration Module (BRM)}
Fig.\,\ref{fig:brm} illustrates the architecture of BRM. At each timestamp, BRM has two responsibilities. First, BRM should provide a background prior for predicting the alpha matte of the current frame. Second, BRM should extract the background information from the current frame and accumulate it to improve the restored background. In our design, BRM achieves these by maintaining (1) a background feature $bgF$ to store the background information and (2) a binary mask $bgM$  to indicate the pixels with restored background.

For the first responsibility, the background feature $bgF^{t-1}$ from the previous timestamp will be regard as the prior for the current frame $I^t$. In practice, we concatenate $bgF^{t-1}$ with $I^{t}$ and the 4x downsampled low-level feature $e_{4\times}^t$ from $\mathcal{S}$ to predict alpha matte $\alpha^t$. For the second responsibility, BRM extracts the background information $bgI^t$ of the current frame by using:
\begin{equation}
    bgI^t = (1 - s_p^t) \otimes e_{4\times}^t,
\end{equation}
where $\otimes$ is pixel-by-pixel multiplication.
$bgI^t$ is then accumulated to $bgF$ under the guided of $bgM$ and $s_p$, as summarized in Alg.\,\ref{alg::BG}. 
In this way, $bgF$, which is continuously delivered and improved in temporal, can provide a strong prior for alpha matte prediction without any user effort or auxiliary inputs. Note that both $bgF$ and $bgM$ are initialized by 0 for the first video frame.

 \begin{figure}[t]
\centering
\begin{tabular}{c@{}c}
\hspace{-2mm}
\includegraphics[width=1.0\linewidth]{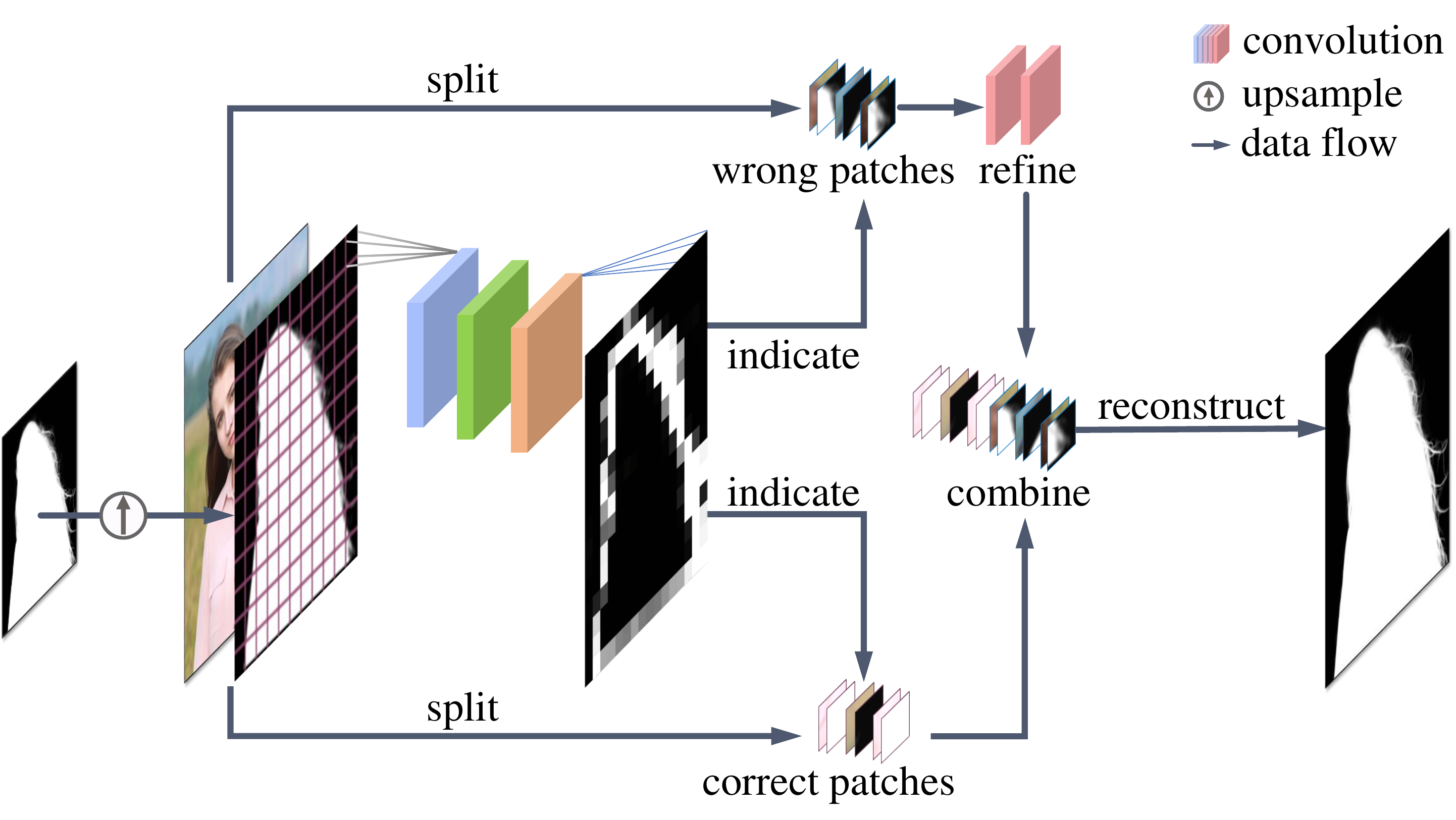}
\ \\
\end{tabular}
{\begin{center}
\vspace{-1mm}
\caption{\small{Illustration of PRM.}}
\label{fig:prm}
\end{center}
}
\vspace{-8mm}
\end{figure}






\subsection{Patch Refinement Module (PRM)}
Although MODNet is carefully designed for reducing inference time, it is still slow when processing images of 1080p or a higher resolution. In order to speed up the matting process of high-resolution images, Background Matting V2 \cite{lin2021real} proposed to first generate a coarse alpha matte from a downsampled input image and then upsample it to the original size for refinement. In the refinement stage, the image with original resolution is cropped into 8x8 patches, and the refinement model only processes the patches with a high error probability. However, we notice that such a strategy has two possible issues when images has 2k or a higher resolution. First, the total number of patches is proportional to image resolution, which means patch selection and cropping will be more time-consuming under a high resolution. Second, as the image resolution increases, the information contained in each 8x8 region will greatly decrease. As a result, artifacts are likely to appear in the edges of refined patches.

In this work, we proposed a new patch refinement module (PRM). As shown in Fig.\,\ref{fig:prm}, different from the size-fixed patch used in Background Matting V2, we divides the input images into a fixed number of patches. Among these patches, the wrong patches will be refined and combined with the other correct patches to form a new alpha matte.

Formally, for an input image of size $(h, w)$, PRM first downsamples it and predict an initial coarse alpha matte. PRM then upsamples the initial coarse alpha matte into the original image size and divides it into $k \times k$ patches of size $(\frac{h}{k}, \frac{w}{k})$. Meanwhile, PRM applies an adaptive pooling layer and two convolutional layers to the initial coarse alpha matte to predict an flaw map (proposed by GCT) of size $(k, k)$, where each pixel value corresponds to the flaw probability of each patch. Finally, PRM only refines the patch whose flaw probability is higher than a predefined threshold $\xi$.
In pracetice, for images with a resolution of 4k and below, we set $k=16$ (the maximum patch size is only $256\times256$). For images with a resolution higher than 4k, we set $k=32$. We set $\xi=0.01$, which constrains only $15\%$ of all patches has a flaw probability higher than $\xi$. This means that PRM can reduce the refinement overhead by $\sim6.5$ times, making it possible for MODNet-V to process high-resolution images.

Our PRM has a consistent time-consuming for patch selection under different input resolutions. In addition, benefit from the larger patch resolution, our PRM can effectively reduce artifacts on fine boundaries (such as the hair regions).
\begin{figure*}[t]
\centering
\begin{tabular}{c@{}c}
\hspace{-3mm}
\includegraphics[width=1\linewidth]{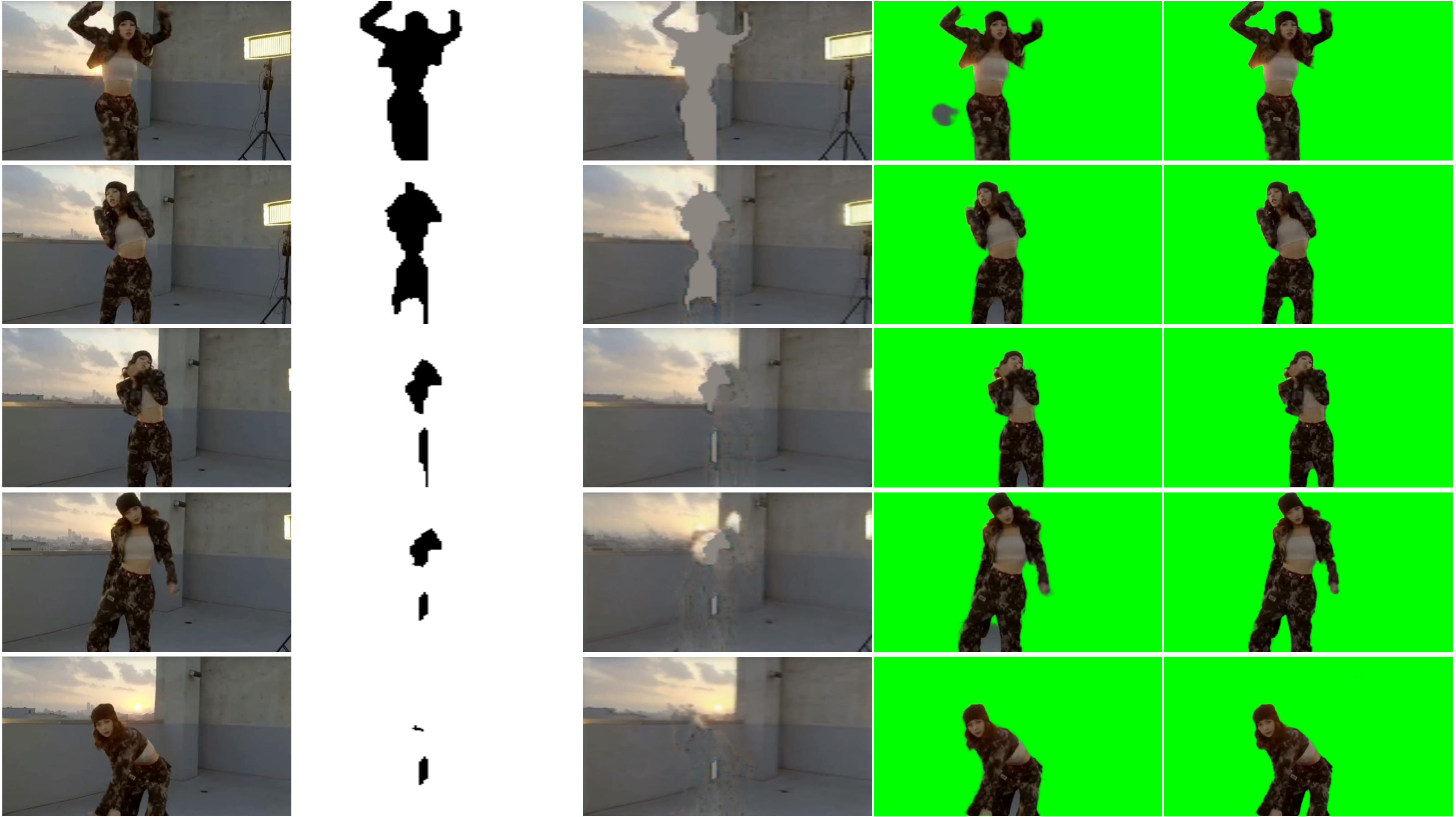}
\ \\
\footnotesize{ \hspace{-2em}(a) Frames \hspace{4em}(b) Background binary masks \hspace{1em}(c) Restored backgrounds \hspace{4em}(d) MODNet \hspace{6em}(e) MODNet-V}\\
\hspace{-3mm}
\end{tabular}
{\begin{center}
\vspace{-4mm}
\caption{\small{Visual Comparison of MODNet and MODNet-V. The proposed BRM restores a background image (c), which helps MODNet-V achieve better and more stable portrait matting results.}}
\label{fig:video1}
\end{center}
}
\vspace{-3mm}
\end{figure*}

\subsection{Training Losses}
We apply loss $\mathcal{L}_{\alpha}$ that is the same as MODNet to learn alpha matte. In addition, we propose $\mathcal{L}_{bg}$ as an explicit constraint to measure the difference between the ground truth background image $bg_g^t$ and the background image $bg_p^t$ predicted from $bgF^t$, as: 
\begin{equation}\label{eq:bg_loss}
\begin{split}
& \mathcal{L}_{bg}= \sum_{t=1}^{N}\gamma \sqrt{\left (bg_p^t - bg_g^t \right )^{2}+\epsilon ^{2}},
\end{split}
\end{equation}
where $\epsilon$ is a small constant value, and $\gamma$ is a binary mask that equals to $4$ in the portrait boundaries otherwise $1$. The purpose of $\gamma$ is to let $\mathcal{L}_{bg}$ focus more on the difficult boundary regions.
We also calculate the constraint $\mathcal{L}_{{\alpha}H}$ between the ground truth high-resolution alpha matte ${\alpha}H_g$ and the predicted high-resolution alpha matte ${\alpha}H_p$ for PRM, as:
\begin{equation}\label{eq:alpha_loss}
\begin{split}
& \mathcal{L}_{{\alpha}H}= \gamma \sqrt{\left ({\alpha}H_p - {\alpha}H_g \right )^{2}+\epsilon^{2}},
\end{split}
\end{equation}
where $\epsilon$ and $\gamma$ has the same meaning as Eq.\,(\ref{eq:bg_loss}).

\section{Experiments}
\subsection{Datasets}
We follow MODNet to train our network on both labeled and unlabeled datasets. \ke{We use the VideoMatte240K dataset~\cite{lin2021real} proposed by Background Matting V2 as the labeled foregrounds and $\sim 1000$ video clips from the internet as the unlabeled data. Note that we also annotate one frame in each video as the labeled data.} 

To prepare a labeled training sample, we randomly take 10 consecutive frames from a video sequence and composite them with a dynamic background sequence that is generated from a single background image by applying translation and affine operations. For each video sequence, we composite it with with $20$ background images, and we finally construct the dataset with $9680$ videos.


\subsection{Training Strategy}
Our MODNet-V applies a two-stage training strategy. 

In the first stage, we train the MODNet component and BRM in an end-to-end manner. To save the GPU memory, in each iteration, we select a random number $T \in [1, 10]$ to indicate the forward frames, and we only calculate the loss of the $1_{th}$ frame and the $T_{th}$ frame for optimization, {\it i.e.}, no gradients are cached and backward among the $[2, T-1]$ frames. As a result, our first training stage can be performed in a single RTX 3090 GPU (24G memory) with a batch size of $16$. We train the first stage for $30$ epochs by using the SGD optimizer. The initial learning rate is $0.01$ and will be decreased by multiplying $0.1$ in the $15_{th}$ and $25_{th}$ epochs. 

In the second stage, we train PRM by using the high-resolution labeled samples only. In this stage, we fix the weights of both the MODNet component and BRM. We use the SGD optimizer with the learning rate of $0.001$ to train this stage for 10 epochs, and we set the batch size to $1$.

\subsection{Performance Comparison}
Table~\ref{tab:experiment} shows the evaluations on a real-world portrait video matting dataset~\cite{wangtiantian21} which contains 19 videos with 711 labeled frames.
%
We can see that the MODNet-V outperforms MODNet in terms of MAD and MSE. In Fig.~\ref{fig:video1}, we visualize the matting results of MODNet and MODNet-V for a video sequence. 

\begin{table}[t]
\setlength{\tabcolsep}{4pt}
\small
\caption{\small{Quantitative results of MODNet and MODNet-V. 
}}
\vspace{1mm}
\centering
\renewcommand{\arraystretch}{1.1}
\begin{tabular}{c||c|c}
\hline
Methods & MAD($10^{-4}$) &MSE($10^{-4}$)\\
\hline \hline
MODNet   &83.30 &34.05          \\
MODNet-V &65.63 &32.39  \\
\hline
\end{tabular}
\vspace{0mm}
\label{tab:experiment}
\end{table}

\section{Conclusion}
In this paper, we presented MODNet-V for portrait video matting. We designed its architecture based on the observation that the background of a video frame can be restored by accumulating the background information from historical frames. In MODNet-V, we proposed a light-weight BRM that performs iteratively with MODNet to construct background as the prior for obtaining better matting results. 
Besides, we introduced a new PRM module that can adapt our method to high-resolution videos. The results on various real-world videos show that the proposed MODNet-V can significantly improve the quality and stability of , compared with previous SOTA like MODNet.

{\small
\bibliographystyle{ieee_fullname}
\bibliography{egbib}
}

\end{document}